\documentclass[11pt,a4paper]{article}
\usepackage[hyperref]{ranlp2025}
\usepackage{times}
\usepackage{latexsym}
\usepackage{booktabs}
\usepackage{siunitx}
\usepackage{amsmath}
\usepackage{amssymb}% http://ctan.org/pkg/amssymb
\usepackage{pifont}% http://ctan.org/pkg/pifont
\newcommand{\cmark}{\ding{51}}%
\newcommand{\xmark}{\ding{55}}%
\usepackage{tabularx}
\usepackage{graphicx}
\usepackage{pifont}
\usepackage{url}

\usepackage{microtype}
\usepackage{listings}
\usepackage{xcolor}
\usepackage{subcaption}
\usepackage{caption}
\usepackage{dblfloatfix}
\usepackage{float}
\usepackage{tcolorbox}
\tcbset{
  aibox/.style={
    width=\columnwidth,
    top=10pt,
    colback=white,
    colframe=black,
    colbacktitle=black,
    center,
  }
}
\newtcolorbox{AIbox}[2][]{aibox,title=#2,#1}  
\lstdefinelanguage{Scala}{
  morekeywords={object,def,if,else,val,var,class,trait,extends,case,Option,Some,None,Int,Double,List},
  sensitive=true,
  morecomment=[l]{//},
  morecomment=[s]{/*}{*/},
  morestring=[b]"
}

\aclfinalcopy % Uncomment this line for the final submission

\title{TypePilot: Leveraging the Scala type system for secure LLM-generated code}

\author{
  {\normalfont \fontsize{10pt}{11pt}\selectfont
  \begin{tabular}[t]{cc}
    \begin{tabular}[t]{@{}l@{}}
      Alexander Sternfeld \\
      Institute of Entrepreneurship \& Management, HES-SO \\
      Le Foyer, Techno-Pôle 1 \\
      Sierre, Switzerland \\
      \texttt{alexander.sternfeld@hevs.ch}
    \end{tabular}
    &
    \begin{tabular}[t]{@{}l@{}}
      Andrei Kucharavy \\
      Institute of Informatics, HES-SO \\
      Techno-Pôle 3 \\
      Sierre, Switzerland \\
      \texttt{andrei.kucharavy@hevs.ch}
    \end{tabular}
  \end{tabular}
  }
  \\
  \\
  {\fontsize{10pt}{11pt}\selectfont
  \begin{tabular}[t]{@{}c@{}}
    Ljiljana Dolamic \\
    Cyber-Defence Campus \\
    armasuisse, Science and Technology \\
    Thun, Switzerland \\
    \texttt{ljiljana.dolamic@armasuisse.ch}
  \end{tabular}
  }
}

\date{}

\begin{document}
\maketitle

\vspace{-8em}

\begin{abstract}
Large language models (LLMs) have shown remarkable proficiency in code generation tasks across various programming languages. However, their outputs often contain subtle but critical vulnerabilities, posing significant risks when deployed in security-sensitive or mission-critical systems. This paper introduces \emph{TypePilot}, an agentic AI framework designed to enhance the security and robustness of LLM-generated code by leveraging strongly typed and verifiable languages, using Scala as a representative example. We evaluate the effectiveness of our approach in two settings: formal verification with the Stainless framework and general-purpose secure code generation. Our experiments with leading open-source LLMs reveal that while direct code generation often fails to enforce safety constraints, just as naive prompting for more secure code, our type-focused agentic pipeline substantially mitigates input validation and injection vulnerabilities. The results demonstrate the potential of structured, type-guided LLM workflows to improve the SotA of the trustworthiness of automated code generation in high-assurance domains.
% \alex{Will have to write a not so abstract abstract}
% ABSTRACT HERE
\end{abstract}

\section{Introduction}

% \alex{Will add citations everywhere}
% \andrei{A general remark - introduction also requires citations. RANLP workshop should be fine, but the citations here are for people from other domains (eg Scala) discovering the field for the first time} 

In recent years, large language models (LLMs) have become powerful tools for assisting in software development, from generating boilerplate code to proposing non-trivial algorithmic implementations ~\cite{wang2023, codex}. Their fluency in natural and programming languages allows developers to interact with them without disrupting their workflow, accelerating the development lifecycle. However, as LLMs are increasingly used to write production code, concerns have emerged about the reliability and security of the generated output. Multiple studies and real-world analyses have shown that LLMs can introduce subtle yet serious vulnerabilities ~\cite{aatk}.

This issue becomes particularly acute in the domain of mission-critical systems—software systems whose failure can lead to catastrophic outcomes, including physical harm, financial loss, major operational disruptions, or loss of life ~\cite{therac}. Such systems are often implemented in strongly typed, safety-oriented programming languages like Coq, Scala, or more recently Rust, where the type system is a central mechanism for enforcing correctness and preventing classes of bugs at run time. Despite these safeguards, vulnerabilities still surface, often due to logical oversights, incorrect assumptions, or abstraction mismatches at boundaries. A well-known example is the 1999 NASA Mars Climate Orbiter failure, where one subsystem produced output in imperial units while another expected metric, leading to the spacecraft's loss due to an undetected discrepancy at the interface between components ~\cite{satellite}. A notable recent example of such a vulnerability in action occurred in January 2023, when a critical FAA system failure, later traced to a corrupted configuration file, led to the temporary grounding of all flights across the United States ~\cite{faa}.

While LLMs are increasingly capable of detecting potential code vulnerabilities, they often fall short in generating robust corrections ~\cite{kulsum2024, pearce2022}. Our work addresses this gap. By focusing on Scala - a widely used language with extensive codebase on GitHub and documentation on StackOverflow ~\cite{redmonk}, we propose TypePilot, an agentic AI approach that not only leverages the detection capabilities of LLMs but actively guides them to exploit the expressiveness of the Scala type system to add safety guarantees. By structuring interactions, TypePilot guides LLMs to generate and refine code that adheres to strict safety and correctness properties.

% \andrei{This part of the of the introduction does not exist in ML papers. However I suggest that you leave the anonimized GitHub link and repeat it once again at the start of the methodology.}
This paper is structured as follows: Section \ref{related_work} describes the related literature, after which Section \ref{methodology} outlines the methodology. Next, the results are presented in Section \ref{results}. Last, Section \ref{conclusion} concludes the paper and provides directions for future research. The code and results related to this paper are publicly available in this \href{https://github.com/fully-anonymized-submission/secure_scala}{Github Repository}.

\section{Related work} \label{related_work}

\begin{figure*}[t]
    \centering
    \includegraphics[width=\linewidth]{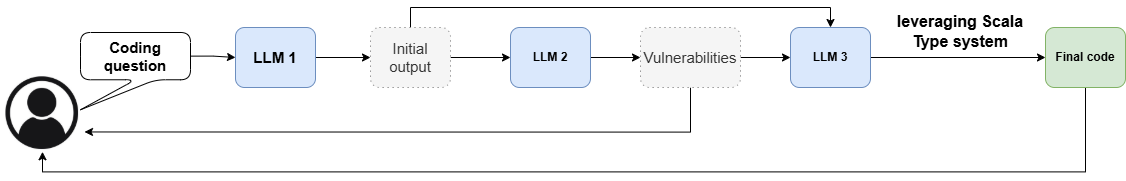}
    \caption{The full pipeline for the generation of code using TypePilot. After the initial generation of the code, the vulnerabilities are detected by a separate instance of the LLM. Then, a final LLM is prompted to leverage the Scala type system to improve the initial code, given the detected vulnerabilities.}
    \label{fig:pipeline}
\end{figure*}

\begin{table*}[b]
\resizebox{\textwidth}{!}{%
\begin{tabular}{@{}lllll@{}}
\toprule
\multicolumn{1}{c}{\textbf{Stainless}}     & \multicolumn{1}{c}{\textbf{}} & \multicolumn{3}{c}{\textbf{General case: type-system rooted vulnerabilities}}                                                                                         \\ \cmidrule(l){3-5} 
                                           &                               & \multicolumn{1}{c}{\textit{Input constraints}} & \multicolumn{1}{c}{\textit{}} & \multicolumn{1}{c}{\textit{Code injection}} \\ \midrule
Calculating a fibonacci number             &                               & Calculating a fibonacci number                 &                               & Greeting a user with HTML                   \\
Calculating the factorial of a number      &                               & Calculating the factorial of a number          &                               & Making a list of comments with HTML         \\
Asserting if list a is a sublist of list b &                               & Calculating a matrix multiplication            &                               & Searching a file using bash                 \\
                                           &                               & Calculating a matrix convolution               &                               & Pinging a host using bash                   \\
                                           &                               &                                                &                               & Creating a redirect URL with HTML           \\ \bottomrule
\end{tabular}%
}
\caption{The test cases used to evaluate the LLMs in each of the settings. The most left column shows the test cases used to evaluate the performance of LLMs in generating code using the Stainless framework. The second and third column show the test cases for the general case looking at type-system rooted vulnerabilities.}
\label{tab:testcases}
\end{table*}

\subsection{LLMs for code generation}
The use of large language models for code generation has grown rapidly, with coding specific models demonstrating impressive capabilities across a wide range of programming languages and tasks. However, several studies have pointed out that these models often produce code that is syntactically correct but semantically flawed or insecure. For instance, ~\citet{aatk} shows that GitHub Copilot produces vulnerabilities in approximately 40\% of test cases based on the top 25 Common Weakness Enumeration list from MITRE. Similarly, ~\citet{khoury2023} show that ChatGPT generates vulnerable code in 16 out of 21 test cases, using a variety of programming languages targeting a diverse set of vulnerabilities. 

% Sandoval et al. (2023) extended this line of work by evaluating the security posture of LLM-generated code in real-world scenarios, finding that vulnerabilities such as SQL injection, path traversal, and race conditions remain common.

There have been attempts to use separate LLMs in combination with sophisticated prompting strategies to patch such vulnerabilities. However, these approaches remain brittle, with models often misunderstanding the root cause or proposing fixes that break functionality. ~\citet{kulsum2024} show that LLMs have difficulty in patching vulnerabilities that are either complex or linked to the project design. Similarly, ~\citet{pearce2022} show that LLMs are not yet able to autonomously patch code vulnerabilities in real-world scenarios. 

Our work builds upon these findings by exploring a different method for mitigating vulnerabilities - leverage the properties of strongly typed coding languages. We use \emph{agentic AI}, where LLMs cooperatively operate as autonomous agents, which has been shown to result in better generations ~\cite{kumar2025, wang2025}. 

% \subsection{Static verification}

% \andrei{There is a paper in ICSE-FORGE on using Vefifast to verify C code generated by GPT4o with a schema similar to Stainless, and it utterly failed: https://conf.researchr.org/details/forge-2025/forge-2025-papers/24/Evaluating-the-Ability-of-GPT-4o-to-Generate-Verifiable-Specifications-in-VeriFast}

% We leverage agentic AI to explicitly use the type system when addressing vulnerabilities and generating code.

% To close the loop between code generation and verification, recent work has explored agentic paradigms, where LLMs operate as autonomous agents capable of planning, tool use, and iteration. Projects such as SWE-agent (Zhu et al., 2023), CodeChain (2024), and Devina (2023) have demonstrated that agent-based LLMs outperform single-shot approaches in tasks like debugging, test-driven development, and long-horizon coding tasks. These agents break down goals, query tools (e.g., linters, tests), interpret feedback, and revise code until a desired objective is met.

% However, few agentic systems explicitly target security or correctness. Most optimize for task completion or functional output, without enforcing safety constraints or using formal feedback mechanisms. Our work extends this paradigm to vulnerability mitigation. We design an agentic system that not only identifies security risks but also uses the Scala type system as a guiding framework for patch generation—revising code until the type checker verifies that safety properties hold. In doing so, we bridge the gap between LLM flexibility and type-theoretic rigor.
\section{Methodology} \label{methodology}
We will now describe the methodology that is used in this research. First, the models that are used in this research are specified. Then, we consider the ability of LLMs to generate code using the formal verification framework Stainless. Last, we consider the general case of type-system rooted vulnerabilities.

\subsection{Model usage} \label{modelusage}
Throughout this research, we use open-source models, with a focus on specialized coding models. Specifically, we used the coding models \texttt{Qwen/Qwen2.5-Coder-32B-Instruct}, \texttt{deepseek-ai/deepseek-coder-33b\-instruct} and \texttt{codellama/CodeLlama\-70b-Instruct-hf}. Additionally, we used the regular conversational models \texttt{meta-llama/Meta-Llama-3-70B}, \texttt{deepseek-ai/DeepSeek-R1-Distill\-Llama-70B} and \texttt{Qwen/Qwen3-32B}. 

% \andrei{Usually we add in the appendix download links for them from HuggingFace to avoid debate over subtle differences in versions.}

\subsection{Stainless}\label{methodstainless}
We first aim to leverage the formal verification framework Stainless \cite{stainless} to improve the robustness of LLM-generated code. Formal verification refers to the use of mathematical methods to prove that a program satisfies certain correctness properties. Stainless is one of the most widely used verification frameworks in Scala, with extensive documentation. Stainless verifies whether Scala code meets user-specified safety properties by attempting to construct proofs over the code. To enable this, the code must explicitly state what is to be proven, and provide the necessary logical structure for the proof, using a subset of Scala tailored for verification.

To this end, we use both zero-shot and two-shot prompting to have a LLM both generate the code and the conditions. Figure \ref{prompt:stainless} displays the prompt that is used in the two-shot prompting setting. The two examples that are given to the LLM are Stainless code for finding the maximum between two values and for returning the size of a list. The exact examples can be found in the \href{https://github.com/fully-anonymized-submission/secure_scala}{Github Repository}. 

As displayed in Table \ref{tab:testcases}, we use three simple tasks for evaluating the LLMs in the context of formal proofs: calculating Fibonacci number n, calculating the factorial of an input and assessing whether list \textit{a} is a sublist of list \textit{b}. The main vulnerability that the generated conditions should prevent are input variables that are invalid, such as a negative input for a factorial function. Additionally, the functions should also be robust to inputs that are too large and may cause an overflow error. 

\begin{figure}[!htb]
\footnotesize
    \begin{AIbox}{Generation of stainless code}
    \textless question\textgreater \space Use the stainless framework to write verifiable scala code for \textcolor{blue}{fewshot example 1} \textless /question\textgreater
    \bigbreak
    \textless answer\textgreater \space \textcolor{blue}{ fewshot answer 1} \textless /answer\textgreater
    \bigbreak
    \textless question\textgreater \space Use the stainless framework to write verifiable scala code for \textcolor{blue}{fewshot example 2 } \textless /question\textgreater
    \bigbreak
    \textless answer\textgreater \space  \textcolor{blue}{fewshot answer 2 } \textless /answer\textgreater
    \bigbreak

    \textless question\textgreater \space  Use the stainless framework to write verifiable scala code for \textcolor{red}{function description} \textless /question\textgreater
    \bigbreak
    \normalsize
    \end{AIbox}
    \caption{Prompt used to generate the Stainless code.}
    \label{prompt:stainless}
\end{figure}

\begin{figure}[!htb]
\footnotesize
    \begin{AIbox}{Code generation in \textit{robust} setting}
    You are a scala code generator. You will be given a task description and you will generate the code for it. The code should start with ```scala and end with ```. Pay attention to the safety and robustness of the code, and leverage the Scala type system - for example ADTs, refined types, traits, sealed traits - where needed to make the code safer. The task is: \textcolor{red}{user input}
    \bigbreak
    \normalsize
    \end{AIbox}
    \caption{Prompt used to generate the code in the robust prompting setting.}
    \label{prompt:robust}
\end{figure}

\begin{figure}[!htb]
\footnotesize
    \begin{AIbox}{Code generation using TypePilot}
    \textbf{Initial code generation}
    \\
    You are a Scala code generator. You will be given a task description and you will generate the code for it. The code should start with ```scala and end with ```. The task is: \textcolor{red}{user input}
    \bigbreak
    \textbf{Vulnerability detection}
    \\
    You will be given a task description and generated code. Your task is to find potential vulnerabilities in the code that could lead to security issues or unexpected behavior. Solely describe the vulnerabilities, do not give me any code. Here is the task: \textcolor{red}{user input} Here is the previous code: \textcolor
    {blue}{initial output}
    \bigbreak
    \textbf{Final code generation}
    \\
    You are a Scala code generator. You will be given a task description, generated code, and vulnerabilities that should be addressed. Your task is to improve the code by using the Scala type system - for example ADTs, refined types, traits, sealed traits - to address the vulnerabilities. The code should start with ```scala and end with ```. Here is the task: \textcolor{red}{user input}. Here is the previous code: \textcolor{blue}{initial output } Here are the vulnerabilities: \textcolor{olive}{vulnerabilities}
    
    \bigbreak
    \normalsize
    \end{AIbox}
    \caption{Prompts used to generate the initial code, the vulnerabilities and the final code in TypePilot. The final prompt guides the LLM to use the Scala type system to make the code more robust.}
    \label{prompt:agentic}
\end{figure}

\subsection{General case: type-system rooted vulnerabilities}\label{methodgeneral}

\begin{table*}[b]
\resizebox{\textwidth}{!}{%
\begin{tabular}{lccccccccccc}
\hline
                                   & \multicolumn{3}{c}{\textbf{Qwen-2.5-Coder (32B)}} &  & \multicolumn{3}{c}{\textbf{CodeLlama (70B)}} &  & \multicolumn{3}{c}{\textbf{Deepseek-coder (33B)}} \\ \hline
                                   & Baseline        & Robust        & TypePilot        &  & Baseline       & Robust      & TypePilot      &  & Baseline        & Robust        & TypePilot        \\ \cline{2-12} 
\textbf{Average age}               &               &               &                   &  &              &             &                 &  &               &               &                   \\
- Correct for regular input        & \cmark        & \cmark        & \cmark            &  & \cmark       & \cmark      & \cmark          &  & \cmark        & \cmark        & \cmark            \\
- Handle empty lists               & \cmark        & \cmark        & \cmark            &  & \cmark       & \cmark      & \cmark          &  & \cmark        & \cmark        & \cmark            \\
- Handle negative ages             & \xmark        & \xmark        & \cmark            &  & \xmark       & \xmark      & \cmark          &  & \xmark        & \xmark        & \xmark            \\
\textbf{}                          &               &               &                   &  &              &             &                 &  &               &               &                   \\
\textbf{Fibonacci number N}        &               &               &                   &  &              &             &                 &  &               &               &                   \\
- Correct for regular input        & \cmark        & \cmark        & \cmark            &  & \cmark       & \cmark      & \cmark          &  & \cmark        & \cmark        & \cmark            \\
- Check for negative N             & \xmark        & \cmark        & \cmark            &  & \xmark       & \xmark      & \xmark          &  & \xmark        & \xmark        & \cmark            \\
- Handles large values of N        & \xmark        & \cmark        & \cmark            &  & \xmark       & \xmark      & \xmark          &  & \xmark        & \cmark        & \cmark            \\
\textbf{}                          &               &               &                   &  &              &             &                 &  &               &               &                   \\
\textbf{Matrix multiplication}     &               &               &                   &  &              &             &                 &  &               &               &                   \\
- Correct for regular input        & \cmark        & \cmark        & \cmark            &  & \cmark       & \cmark      & \cmark          &  & \cmark        & \cmark        & \cmark            \\
- Check for empty matrices         & \cmark        & \xmark        & \cmark            &  & \xmark       & \xmark      & \xmark          &  & \xmark        & \xmark        & \xmark            \\
- Check for dimension matching     & \cmark        & \cmark        & \cmark            &  & \cmark       & \xmark      & \cmark          &  & \xmark        & \cmark        & \cmark            \\
\textbf{}                          &               &               &                   &  &              &             &                 &  &               &               &                   \\
\textbf{Matrix convolution}        &               &               &                   &  &              &             &                 &  &               &               &                   \\
- Correct for square matrix input  & \cmark        & \cmark        & \cmark            &  & \xmark       & \xmark      & \xmark          &  & \cmark        & \cmark        & \cmark            \\
- Correct for regular matrix input & \cmark        & \cmark        & \cmark            &  & \xmark       & \xmark      & \xmark          &  & \cmark        & \cmark        & \cmark            \\
- Handles rectangular kernels      & \xmark        & \xmark        & \cmark            &  & \xmark       & \xmark      & \cmark          &  & \xmark        & \xmark        & \xmark            \\
- Checks for empty kernel          & \xmark        & \cmark        & \cmark            &  & \xmark       & \xmark      & \cmark          &  & \xmark        & \xmark        & \cmark            \\
- Checks for empty matrix          & \xmark        & \cmark        & \cmark            &  & \xmark       & \xmark      & \cmark          &  & \xmark        & \xmark        & \cmark            \\
- Handles even sized kernels       & \xmark        & \xmark        & \cmark            &  & \xmark       & \xmark      & \xmark          &  & \xmark        & \xmark        & \xmark            \\ \hline
\end{tabular}%
}
\caption{Manual evaluation of the generated code regarding input constraints. For each case, \cmark \space indicates that the code is robust to the vulnerability, whereas \xmark \space indicates that the code is not robust to the vulnerability.}
\label{tab:input}
\end{table*}

\begin{table*}[t]
\resizebox{\textwidth}{!}{%
\begin{tabular}{lccccccccccc}
\hline
                            & \multicolumn{3}{c}{\textbf{Qwen-2.5-Coder (32B)}} &  & \multicolumn{3}{c}{\textbf{CodeLlama (70B)}} &  & \multicolumn{3}{c}{\textbf{Deepseek-coder (33B)}} \\ \hline
                            & Baseline    & Robust    & TypePilot    &  & Baseline   & Robust  & TypePilot  &  & Baseline    & Robust    & TypePilot    \\ \cline{2-12} 
\textbf{}                   &             &                     &               &  &            &                   &             &  &             &                     &               \\
\textbf{HTML greeting}      &             &                     &               &  &            &                   &             &  &             &                     &               \\
Correctness and compilation & \cmark      & \cmark              & \cmark        &  & \cmark     & \cmark            & \cmark      &  & \cmark      & \cmark              & \xmark        \\
Robust to injection         & \xmark      & $\sim$                & \cmark        &  & \xmark     & \cmark            & \cmark      &  & \xmark      & \cmark              & \cmark        \\
\textbf{}                   &             &                     &               &  &            &                   &             &  &             &                     &               \\
\textbf{HTML comments}      &             &                     &               &  &            &                   &             &  &             &                     &               \\
Correctness and compilation & \cmark      & $\sim$                & \cmark        &  & \cmark     & \xmark            & $\sim$        &  & \cmark      & \cmark              & \cmark        \\
Robust to injection         & \xmark      & \cmark              & \cmark        &  & \xmark     & \xmark            & \cmark      &  & \xmark      & $\sim$                & \cmark        \\
                            &             &                     &               &  &            &                   &             &  &             &                     &               \\
\textbf{Bash file search}   &             &                     &               &  &            &                   &             &  &             &                     &               \\
Correctness and compilation & \cmark      & \cmark              & \cmark        &  & \cmark     & \cmark            & \cmark      &  & \cmark      & \cmark              & \cmark        \\
Robust to injection         & \xmark      & \xmark              & \cmark        &  & \xmark     & \xmark            & \cmark      &  & \xmark      & \xmark              & \cmark        \\
                            &             &                     &               &  &            &                   &             &  &             &                     &               \\
\textbf{Bash host ping}     &             &                     &               &  &            &                   &             &  &             &                     &               \\
Correctness and compilation & \cmark      & \cmark              & \cmark        &  & \cmark     & \xmark            & \cmark      &  & \cmark      & \cmark              & \cmark        \\
Robust to injection         & \cmark      & \cmark              & \cmark        &  & \xmark     & \xmark            & \cmark      &  & \xmark      & \cmark              & \cmark        \\
                            &             &                     &               &  &            &                   &             &  &             &                     &               \\
\textbf{URL redirect}       &             &                     &               &  &            &                   &             &  &             &                     &               \\
Correctness and compilation & \cmark      & \cmark              & \cmark        &  & \cmark     & \cmark            & \cmark      &  & \cmark      & \cmark              & \cmark        \\
Robust to injection         & \xmark      & $\sim$                & $\sim$          &  & \xmark     & \cmark            & \cmark      &  & \xmark      & \xmark              & \cmark        \\ \hline
\end{tabular}%
}
\caption{Manual evaluation of the generated code regarding code injection. For each case, \cmark \space indicates that the code is robust to the vulnerability, whereas \xmark \space indicates that the code is not robust to the vulnerability.}
\label{tab:injection}
\end{table*}

As Stainless targets a niche subset of Scala applications, we also consider a more general setting. Specifically, we focus on two vulnerability categories: insufficient input constraints and injection attacks. In particular, we examine HTML, Bash, and URL injections—common security risks in back-end web development, especially when handling user inputs through web forms. The specific test cases are shown in Table \ref{tab:testcases}. To assess the performance of LLMs on these tasks, we consider the following settings:

\begin{itemize}
    \item \textbf{Baseline: } directly prompting a LLM to generate the code
    \item \textbf{Robust prompting: }directly prompting a LLM to generate the code, while emphasizing that the LLM should leverage the Scala type system to make the code robust to potential vulnerabilities.
    \item \textbf{TypePilot: }use the agentic AI framework as displayed in Figure \ref{fig:pipeline} to generate the code. After prompting a first LLM to generate the initial code, we ask a second LLM to detect the vulnerabilities in this code. We then ask a third instance of the LLM to improve the initial code using the Scala type system, to make it robust to the detected vulnerabilities. 
\end{itemize}

The prompt that is used in the robust generation setting can be found in Figure \ref{prompt:robust}. Similarly, Figure \ref{prompt:agentic} shows the prompts that are used with TypePilot. In the baseline setting, we use the same prompt that is used for the initial code generation in TypePilot. For each of the models described in Section \ref{modelusage} we run each of the settings. 

\subsection{Comparison to existing work} \label{comparison}

Research on secure code generation using large language models (LLMs) remains limited, despite growing concerns about vulnerabilities in automatically generated code. A recent survey by ~\citet{dai2025} highlights that most current approaches rely heavily on training data or static analysis tools, restricting their generalizability. Methods such as SafeCoder ~\cite{safecoder} and SVEN ~\cite{sven} fine-tune LLMs with curated secure code datasets, and are thus inherently dependent on the availability and quality of specialized training corpora. Moreover, the fine-tuned LLMs do not generalize well to unseen vulnerabilities or programming languages. Similarly, PromSec ~\cite{promsec} optimizes prompts through static analyzers, but also relies on labeled data and an external code-specific vulnerability scanner.

In contrast, our approach does not rely on task-specific training data or external static analyzers. Instead, it leverages the expressive power of strongly typed languages to enforce security constraints directly in the generated code. Because most existing methods depend on curated datasets or vulnerability scanners (as discussed above), there are few established baselines tailored to strongly typed languages like Scala or Rust. Given this gap, it is most appropriate to compare our method against prompting-based baselines in addition to the base model. Following ~\citet{baxbench}, we include a baseline where the model is given a general security reminder, which we call \emph{robust prompting}. We also evaluate against \emph{Self-Planning}, a coding-specific prompting strategy introduced by ~\citet{selfplanning}. Self-Planning is a two-stage prompting framework in which the LLM first generates a high-level plan for the coding task, after which it implements the plan in code.

\section{Results} \label{results}

\subsection{Stainless}
% Examples: casting bigInt to int, using println, using List.sliding, using for-all with index aware lambdas (.forall((x, i) => ...))
% Let us start by describing the results using the Stainless framework. \andrei{not really needed in scientific writing}
In general, we see that none of the models is capable of consistently generating Stainless code that correctly compiles. Upon manual inspection, we found two main failure modes across all models. First, each of the LLMs regularly uses concepts that are present in Scala but not available in Stainless. As Stainless is a verification framework targeting a restricted subset of Scala, many features of full Scala—such as certain standard library functions—are unavailable. To illustrate, in the generated code from \texttt{Qwen/Qwen3-32B} for the verification of a sublist relation, the function \texttt{List.sliding} is used. However, the sliding operation is not defined for Stainless \texttt{List} objects. Similarly, in a generated code snippet the operation \texttt{println} was used, which is not available in Stainless. Second, the generated code often contains syntax errors. Whereas syntax errors could be resolved relatively easily by users, the usage of Scala components in Stainless is not trivially repaired. We hypothesize that the lack of performance is caused by a lack of training data related to Stainless, given that it is a niche framework. This observation is consistent with findings from other domains, for example, \citet{fan2025} found that LLMs struggle to generate verifiable specifications using the VeriFast verification framework for C, despite preserving functional behavior. In appendix B, we provide a notable instance in which the generation avoids formal verification by using \texttt{@library} annotations.

% One notable exception where the generated code did compile involved the use of \texttt{@library} (and in other cases, \texttt{@extern}) annotations. These annotations are used to mark functions that should be excluded from formal verification by Stainless, as they have been proven externally. In essence, they allow code to compile by bypassing the verification checks entirely. While this results in syntactically valid and compilable code, it undermines the core purpose of using Stainless — to provide formal guarantees. The use of these annotations masks verification failures rather than solving them, making the code appear correct when it has not actually been verified. This introduces a serious risk: a user may assume the model has correctly verified the code, when in reality it has simply been excluded from verification. 

\begin{figure*}[b]
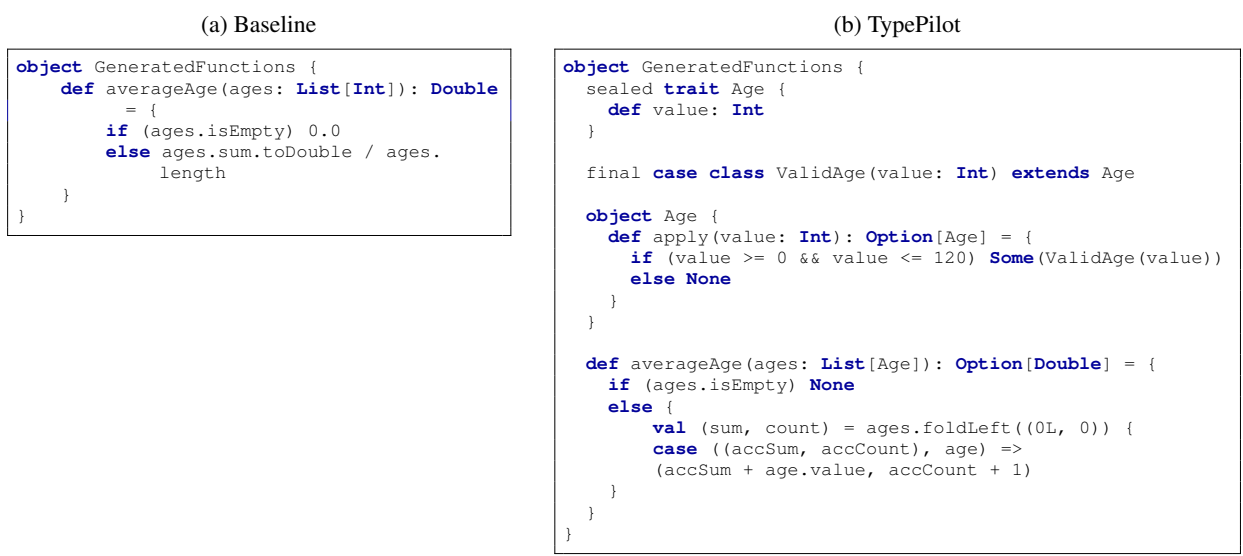

\centering
\begin{subfigure}[t]{0.4\textwidth}
\centering
\caption{Baseline}
\begin{minipage}{\linewidth}
\begin{lstlisting}[language=Scala]
object GeneratedFunctions {
    def averageAge(ages: List[Int]): Double = {
        if (ages.isEmpty) 0.0
        else ages.sum.toDouble / ages.length
    }
}
\end{lstlisting}
\end{minipage}
\end{subfigure}
\hfill
\begin{subfigure}[t]{0.55\textwidth}
\centering
\caption{TypePilot}
\begin{minipage}{\linewidth}
\begin{lstlisting}[language=Scala]
object GeneratedFunctions {
  sealed trait Age {
    def value: Int
  }

  final case class ValidAge(value: Int) extends Age

  object Age {
    def apply(value: Int): Option[Age] = {
      if (value >= 0 && value <= 120) Some(ValidAge(value))
      else None
    }
  }

  def averageAge(ages: List[Age]): Option[Double] = {
    if (ages.isEmpty) None
    else {
        val (sum, count) = ages.foldLeft((0L, 0)) { 
        case ((accSum, accCount), age) =>
        (accSum + age.value, accCount + 1)
    }
  }
}
\end{lstlisting}
\end{minipage}
\end{subfigure}
\caption{Comparison of baseline and TypePilot average age function generations from \texttt{Qwen-2.5-Coder (32B)}}
\label{fig:average-age-comparison}
\end{figure*}

% \begin{figure*}[b]
% \centering
% \nolinenumbers
% \begin{minipage}{0.85\linewidth}
% \begin{minted}[fontsize=\scriptsize]{scala}
% @extern
% def correctness(n: BigInt): Boolean = {
%   require(n >= 0)
%   n match {
%     case 0 => fib(n) == 0
%     case 1 => fib(n) == 1
%     case _ => fib(n) == fib(n - 1) + fib(n - 2)
%   }
% }.holds

% @extern
% def everyThirdIsEven(n: BigInt): Boolean = {
%   require(n >= 0)
%   (n % 3 == 0) implies (fib(n) % 2 == 0)
% }.holds
% \end{minted}
% \end{minipage}
% \caption{Verification conditions for Fibonacci correctness and parity properties.}
% \label{fig:fib-verification}
% \end{figure*}

\subsection{General setting}
Given that LLMs are not able to write compilable Stainless code, we shift our attention to a more general Scala setting, as described in Section \ref{methodgeneral}. We consider two types of vulnerabilities: insufficient input constraints and code injection. The generated code is available in the anonymized repository. 

\subsubsection{Input constraints}
Table \ref{tab:input} shows the results for each of the test cases for each of the models. For each of the models, \xmark \space indicates that the resulting code was not robust to the indicated vulnerability. The results show that in the baseline setting, the models are capable of generating functions that provide the correct output in a normal setting. However, the models are not capable of handling edge cases correctly. To illustrate, none of the models can correctly handle negative ages or a negative input to a Fibonacci function. We see that in the \emph{robust} setting, models perform slightly better, and tend to be robust to some of the vulnerabilities. However, for none of the models the code is fully robust. With TypePilot, we obtain the best performance, with models generally being robust to most vulnerabilities related to input constraints.  

When comparing the models, we observe that \texttt{Qwen-2.5-Coder (32B)} performs the best, passing all our checks when using TypePilot. In contrast, \texttt{CodeLlama (70B)} does not perform well, remaining vulnerable to a number of cases in each of the settings, highlighting the importance of study of specific code-generating LLM models.

\subsubsection{Code injection}
The second type of vulnerability we consider is code injection. Table \ref{tab:injection} displays the results for each of the models, where \xmark \space indicates that the code is vulnerable to injection, \cmark \space indicates that code is robust to injection, and $\sim$ indicates that the code is partially robust to injection.  The results show that in the baseline setting virtually all generated code is vulnerable to code injection. Robust prompting improves the performance, resulting in fewer vulnerabilities. As before, TypePilot achieves the best performance, with robust code generations in almost all settings.

\begin{figure*}[b]
    \centering
    \includegraphics[width=0.8\linewidth]{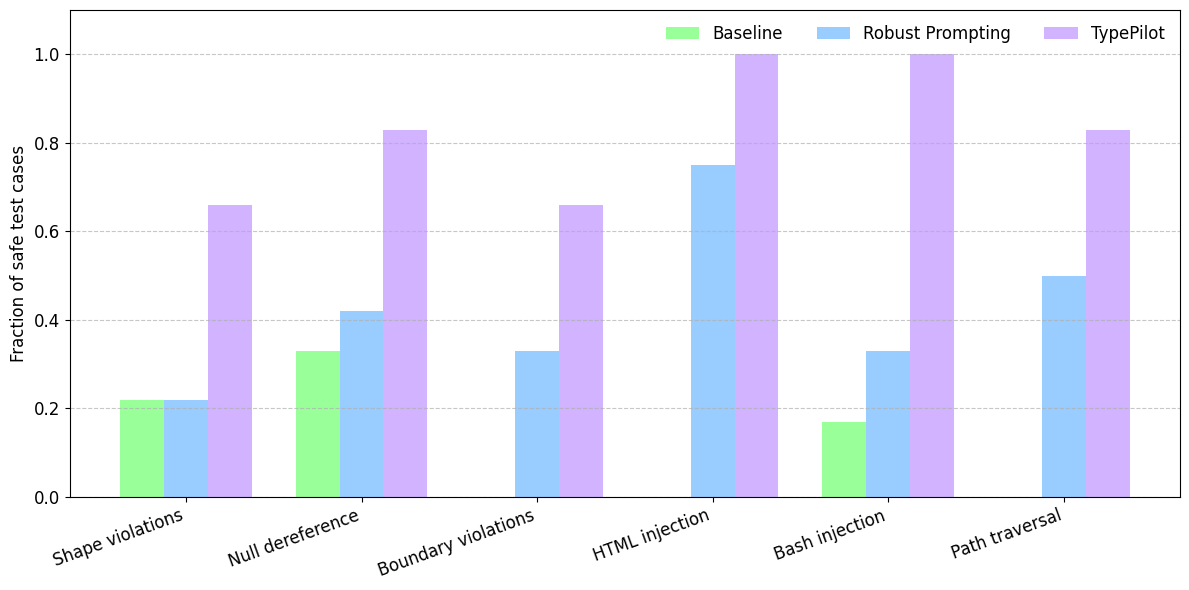}
    \caption{Fraction of secure code generations across vulnerability categories for each of the methods (baseline prompting, robust prompting, TypePilot). Results are averaged over all evaluated LLMs. Lower bars indicate a higher frequency of vulnerabilities; higher bars indicate safer generations.}
    \label{fig:vuln_analysis}
\end{figure*}

\subsubsection{Usage of the Scala type system}
In the new framework, the Scala type system is used as a central tool to guide the generation of secure code. By prompting LLMs to leverage features such as sealed traits, smart constructors, and refined return types, we enable the generation of programs that encode correctness directly into their type signatures. This stands in contrast to baseline generations, which operate on unconstrained primitives and rely on ad hoc runtime logic to handle edge cases and errors.

Figures \ref{fig:average-age-comparison} shows an examples of code generated in the baseline and in the agentic AI framework, for the same test case and model. Figure \ref{fig:average-age-comparison} shows that in the baseline version, the \texttt{averageAge} function takes a \texttt{List[Int]} and performs a division after checking for emptiness. While this implementation is syntactically valid, it permits semantically invalid inputs—such as negative ages or values far outside a realistic human range—and silently defaults to returning 0.0 when the input list is empty. In contrast, the enhanced version defines a sealed trait \texttt{Age} and a case class \texttt{ValidAge}, with a smart constructor in the \texttt{Age} companion object that enforces domain-specific constraints: only values between 0 and 120 are permitted. The \texttt{averageAge} function now accepts a list of validated \texttt{Age} values and returns an \texttt{Option[Double]}, making both the domain invariants and the possibility of undefined results (e.g., empty lists) explicit at the type level. This design ensures that all inputs have been prevalidated before the function executes, reducing the likelihood of subtle logic bugs and enabling safer composition in larger systems. A second example related to generating a function to search for files using bash is discussed in appendix C.

In TypePilot, the Scala type system is used not merely to enforce syntactic correctness but to encode domain abstractions rules, constrain behavior, and make failure modes explicit. By doing so, it transforms what would otherwise be runtime checks and ad hoc validations into statically enforced contracts. This shift leads to code that is more robust, more predictable, and better aligned with the principles of secure and maintainable software design. In the context of LLM-generated code, these benefits are particularly important, as they offer a principled way to guard against common pitfalls and encourage safer defaults during generation.

\subsection{Vulnerability Analysis}
We performed a post-hoc vulnerability analysis by categorizing the vulnerabilities observed in each test case. These categories include input constraint issues (shape violations, null dereferences and boundary violations) and code injection risks (HTML injection, bash injection and path traversal). For each method, we calculated the fraction of secure outputs and averaged the results across the three LLMs, which is displayed in Figure \ref{fig:vuln_analysis}.

For input constraints, robust prompting offered limited improvements over the baseline, particularly for shape violations and null dereferences. It often inserted assertions but did not systematically enforce data structure correctness. TypePilot reduced these errors more effectively, as the presence of type specifications led the models to generate code structured around expected data formats rather than relying on runtime checks.

For code injection, TypePilot also lowered vulnerability rates, especially for bash injections where robust prompting typically altered command structure without validating input. Results varied between models: Qwen-2.5-Coder (32B) and Deepseek-Coder (33B) generally applied the type system consistently, while CodeLlama (70B) sometimes attempted to handle vulnerabilities outside the type framework. In some cases, type constraints were only partially used, such as defining a type for an output value but not for the input values.

Appendix D analyzes attention weights across the three methods, showing that TypePilot places greater emphasis on key safety terms during code generation than robust prompting.

\subsection{Comparison to Self-Planning Code Generation}

As an additional validation, we compared TypePilot to the Self-Planning prompting framework, as discussed in Section \ref{comparison}. In the Self-Planning framework, the model is first asked to outline a plan for solving the task. Afterwards, it is asked to write the code by executing the plan, and it is explicitly instructed to consider safety and security aspects before writing code. Overall, TypePilot outperforms self-planning for both the input constraint and code injection tasks. The difference is largest for Qwen-2.5-Coder (32B), which more reliably adheres to the type system instructions in TypePilot, resulting in fewer shape and null-handling issues compared to the Self-Planning setup.

Manual inspection of the Self-Planning outputs reveals that, despite explicit prompts to account for vulnerabilities during the planning and implementation stages, models frequently overlook or under-address these concerns. The generated plans may mention security considerations in abstract terms but rarely translate them into concrete, protective measures in the final code. These findings suggest that simply instructing the model to “think about safety” is insufficient: introducing a structured intermediate step, such as TypePilot’s type-enforced specification phase, is more effective in steering the model toward safer code generation.

\begin{figure}
    \centering
    \includegraphics[width=\linewidth]{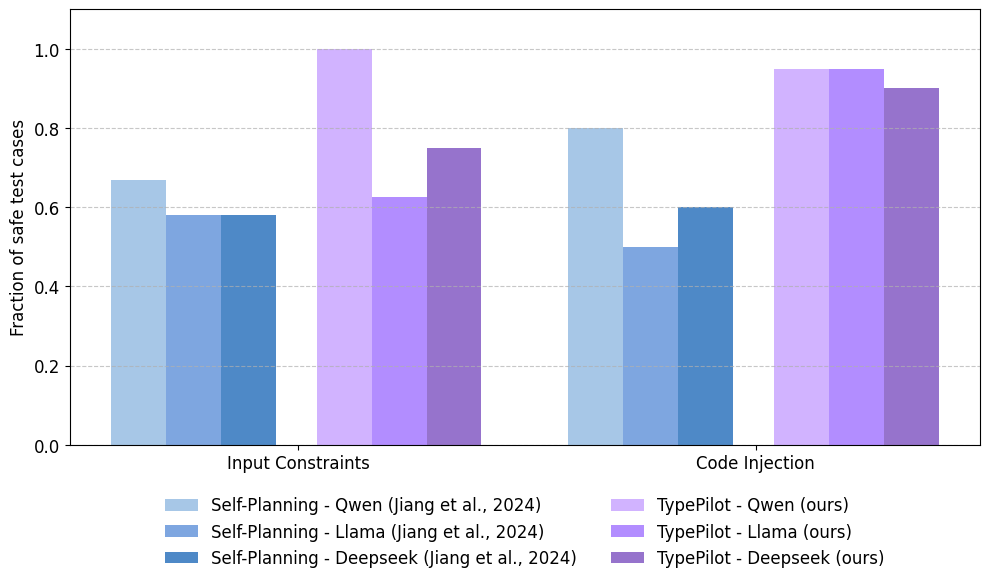}
    \caption{Comparison of the secure code generation methods TypePilot (ours) and Self-Planning, as introduced by \citet{selfplanning}.}
    \label{fig:selfplanning}
\end{figure}

\section{Scaling}

The primary goal of this work is to show that leveraging the type system in strongly typed languages can substantially mitigate vulnerabilities in LLM-generated code. While our experiments focus on relatively simple test cases, practical applications often involve larger, interconnected codebases with complex object hierarchies. Scaling our framework to such scenarios presents new challenges, primarily related to context management and dependency reasoning across multiple files and modules.

One promising direction is the development of a hybrid, object-aware prompting system. In this approach, metadata about each relevant object, including its types and invariants, is provided to the LLM prior to generation. This structured context could enable the model to reason more accurately about type interactions and enforce security constraints across function boundaries. Additionally, integrating lightweight symbolic reasoning or type inference engines could help LLMs maintain global consistency in larger projects, further reducing the risk of injection attacks and logical errors. 

\section{Conclusion} \label{conclusion}

In this work, we aim to improve the security of LLM generated mission-critical code, focusing on the Scala strongly typed language.
% Scala code.
As Scala is routinely used in mission-critical software and engineers are increasingly often using LLMs to code, it is essential to ensure that the generated code is free of vulnerabilities. We first show that LLMs are not able to autonomously use the static verification tool Stainless. Therefore, we develop a more general agentic AI framework that structures multi-step interactions between LLMs for code generation. By leveraging the Scala type system, we significantly improve the quality and safety of generated code. Crucially, this approach transforms type systems from passive compile-time enforcers into active agents of code safety. We study two different classes of vulnerabilities, input constraints and code injection, and show that in both cases our framework improves code safety over a baseline and zero-shot robust prompting setting. We use the rigidity of the Scala type system to compensate for the inconsistencies picked up from the training code by LLMs, which in turn allow an easier interface to access the power of the Scala type system.

We conclude by suggesting two directions for future research. First, future work should test the framework’s capabilities in more complex codebases. While this study provided a proof of concept using simple test cases, real-world software tends to be more complex, so validating our approach in these environments is important to assess its effectiveness. Second, deploying the framework in an active development setting would allow engineers to use it in their daily work and provide valuable feedback. This real-world input can guide further improvements and help tailor the framework to better meet the needs of software teams.

\bibliographystyle{acl_natbib}
\bibliography{ranlp2025}

\begin{figure*}[b]
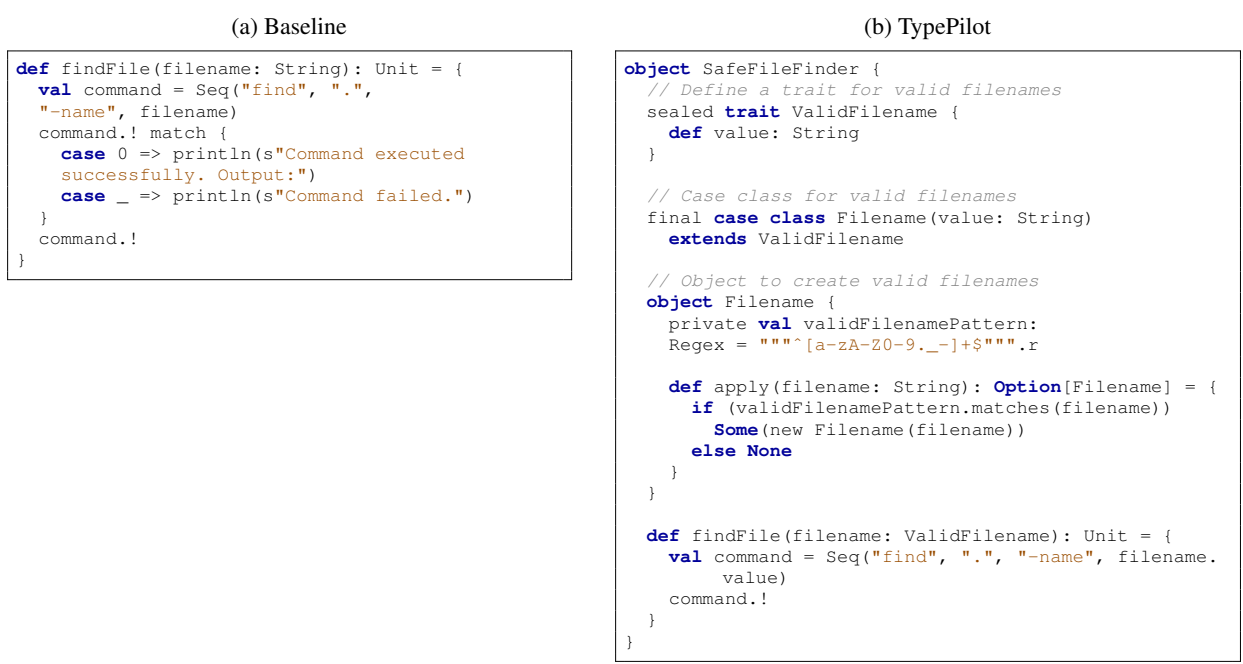

\centering

\begin{subfigure}[t]{0.45\textwidth}
\centering
\caption{Baseline}
\begin{minipage}{\linewidth}
\begin{lstlisting}[language=Scala]
def findFile(filename: String): Unit = {
  val command = Seq("find", ".", 
  "-name", filename)
  command.! match {
    case 0 => println(s"Command executed 
    successfully. Output:")
    case _ => println(s"Command failed.")
  }
  command.!
}
\end{lstlisting}
\end{minipage}
\end{subfigure}
\hfill
\begin{subfigure}[t]{0.5\textwidth}
\centering
\caption{TypePilot}
\begin{minipage}{\linewidth}
\begin{lstlisting}[language=Scala]
object SafeFileFinder {
  // Define a trait for valid filenames
  sealed trait ValidFilename {
    def value: String
  }

  // Case class for valid filenames
  final case class Filename(value: String) 
    extends ValidFilename

  // Object to create valid filenames
  object Filename {
    private val validFilenamePattern: 
    Regex = """^[a-zA-Z0-9._-]+$""".r

    def apply(filename: String): Option[Filename] = {
      if (validFilenamePattern.matches(filename)) 
        Some(new Filename(filename))
      else None
    }
  }

  def findFile(filename: ValidFilename): Unit = {
    val command = Seq("find", ".", "-name", filename.value)
    command.!
  }
}
\end{lstlisting}
\end{minipage}
\end{subfigure}

\caption{Comparison of baseline and TypePilot implementations of file-finding functionality.}
\label{fig:file-find-comparison}
\end{figure*}

% ---------------- FIGURE 2 ----------------
\begin{figure*}[t]
\centering
\begin{lstlisting}[language=Scala,frame=lines,basicstyle=\ttfamily\small]
import stainless.annotation._
import stainless.lang._
import stainless.collection._

object Factorial {

    @library
    def factorial(n: BigInt): BigInt = {
        decreases(n)
        if (n == 0) BigInt(1)
        else n * factorial(n - 1)
    }

    @library
    def factorialCorrectness(n: BigInt): Boolean = {
        def factorialInduction(k: BigInt): Boolean = {
            decreases(k)
            if (k == 0) true
            else {
                val fac = factorial(k)
                fac == k * factorial(k - 1) && factorialInduction(k - 1)
            }
        }
        factorialInduction(n)
    }

    def factorialCorrect(n: BigInt): Boolean = {
        // Checks that the factorial is correct for all n.
        require(n >= 0)
        factorialCorrectness(n)
    }

    def factorialLteN(n: BigInt): Boolean = {
        // Checks that the factorial is always larger or equal to n for all n.
        require(n >= 0)
        factorial(n) >= n
    }

}
\end{lstlisting}
\caption{Generated Stainless code for computing the factorial of a given number with correctness checks. Due to the \texttt{@library} annotations the code silently passes, without actual verification, leading to potential security risks.}
\label{fig:factorial-stainless}
\end{figure*}
\section*{A. Computational resources and environmental impact}

Training and evaluating large language models (LLMs), especially in an agentic framework, comes with significant computational and environmental costs. For the experiments described in this paper, we used a computing setup consisting of 4 NVIDIA A100 GPUs (40GB each). Each full agentic AI pipeline evaluation, which involved three separate LLM calls (initial generation, vulnerability detection, and improved generation), took approximately 45 seconds per test case, depending on the model size and prompt complexity.

Given that one A100 GPU under load emits a maximum of 400W, and considering the CO$_2$ emissions per kilowatt-hour in the region we are located in to be $\SI{38.30} {\gram\text{CO}_{2}\text{eq/kWh}}$, we estimate the carbon footprint per agentic AI test case to be approximately 0.08 g $\text{CO}_\text{2}$ eq. For the baseline and robust setting, each test case results in approximately 0.03 g $\text{CO}_\text{2}$ eq. Taking into account all evaluations, the combined total emissions of this project are 3.76 g $\text{CO}_\text{2}$ eq.

We emphasize that our framework is not intended for deployment at industrial or web-scale. Rather, the motivation for this work lies in high-assurance or mission-critical software development contexts where correctness and safety outweigh considerations of latency or throughput. In domains such as aviation control software, medical devices, or nuclear infrastructure, even a small gain in robustness may justify the computational overhead. Furthermore, this framework serves as a proof-of-concept for incorporating type-guided LLM reasoning into secure code generation, which could eventually be optimized or distilled into more efficient forms.

To promote reproducibility while minimizing redundant computational cost, we have made our prompts, outputs, and testing infrastructure publicly available, enabling researchers to reuse or adapt our setup. The code is available through this \href{https://github.com/fully-anonymized-submission/secure_scala}{Github link}.

\begin{figure*}[t]
  \centering

  % Left stacked images (in a minipage to fix alignment)
  \begin{subfigure}[t]{0.48\textwidth}
    \vspace{0pt} % Force top alignment
    \begin{minipage}[t]{\linewidth}
      \begin{subfigure}[t]{\linewidth}
        \includegraphics[width=\linewidth]{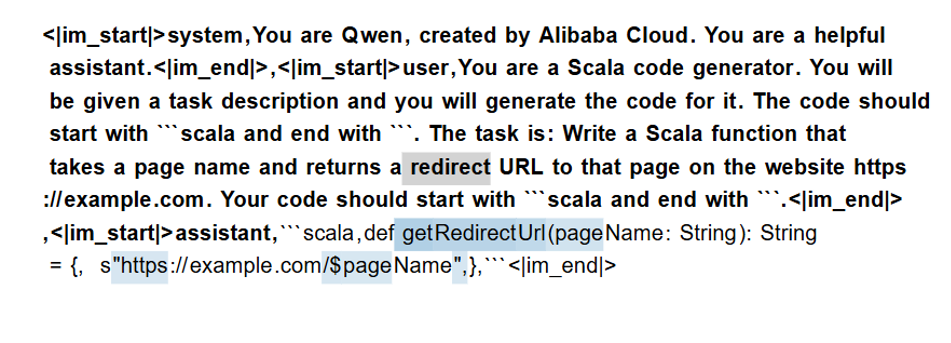}
        \caption{Baseline}
        \label{fig:baseline}
      \end{subfigure}

      \vspace{0.02\textwidth} % Space between the stacked images

      \begin{subfigure}[t]{\linewidth}
        \includegraphics[width=\linewidth]{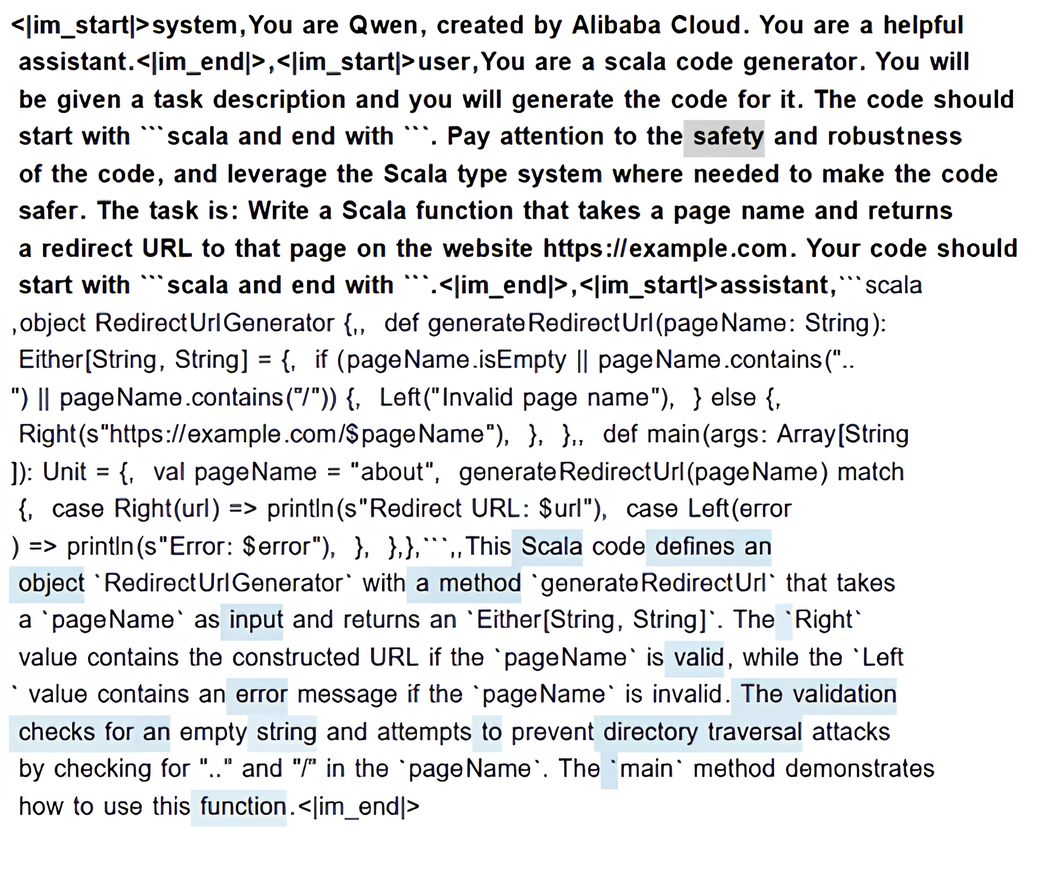}
        \caption{Robust}
        \label{fig:robust}
      \end{subfigure}
    \end{minipage}
  \end{subfigure}
  \hfill
  % Right single image
  \begin{subfigure}[t]{0.48\textwidth}
    \vspace{0pt} % Force top alignment
    \includegraphics[width=\linewidth]{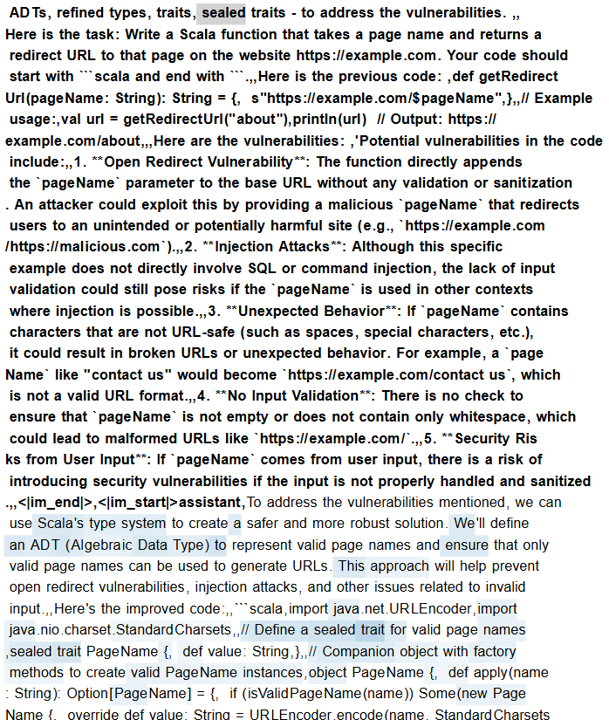}
    \caption{TypePilot}
    \label{fig:agentic}
  \end{subfigure}

  \caption{Comparison for the generation of Scala code for a URL redirection function between the baseline, robust and TypePilot approaches. The source token is highlighted in grey, and the tokens that have non-zero attention weights are marked blue. A darker color blue reflects a higher attention weight.}
  \label{fig:comparison}
\end{figure*}

\section*{B. Example generation in Stainless}
 This appendix belongs to Section 4.1 of the main paper. Here we present one notable exception of a generation of code in Stainless where the generated code did compile. This is shown in Figure \ref{fig:factorial-stainless} and involved the use of \texttt{@library} annotations. These annotations are used to mark functions that should be excluded from formal verification by Stainless, as they have been proven externally. In essence, they allow code to compile by bypassing the verification checks entirely. While this results in syntactically valid and compilable code, it undermines the core purpose of using Stainless — to provide formal guarantees. The use of these annotations masks verification failures rather than solving them, making the code appear correct when it has not actually been verified. This introduces a serious risk: a user may assume the model has correctly verified the code, when in reality it has simply been excluded from verification.

\section*{C. Comparison of baseline and TypePilot}
Here, we provide a second comparison between TypePilot and the baseline, similar to the example in Figure 5 in the paper. This example involves generating a function to search for files using the find command. The baseline implementation constructs the command by interpolating a user-supplied String, which introduces a risk of command injection if the string contains shell metacharacters or other malicious content. This also exposes the system to path traversal vulnerabilities, where attackers may input sequences like "../.." to escape intended directories. This issue has been at the core of high-profile security incidents, for instance a recent vulnerability in the Ivanti Endpoint Manager ~\cite{ivanti}. TypePilot redefines the interface around a sealed trait \texttt{ValidFilename} and a case class \texttt{Filename}, whose creation is guarded by a regular expression enforcing a strict naming convention. Only inputs that match this pattern are allowed to construct a \texttt{Filename} instance, and the \texttt{findFile} function requires this type, thereby rejecting unsafe inputs at compile time. This approach ensures that potentially harmful inputs are never passed into shell command contexts without validation, effectively eliminating an entire class of vulnerabilities by design.

\section*{D. Attention Mechanisms}

The analysis of attention weights is a well-established method for understanding how encoder models interpret and process short-sequence input data, with BertViz being a prominent example. Through the visualization of attention weights, we can obtain insights into which input tokens had the most influence on the prediction of each output token. While BertViz was designed primarily for encoder models and short input contexts, its extension to autoregressive models such as large language models remains an area of ongoing exploration. Here, we aim to get preliminary insights into the functionality of TypePilot through the attention weights. 

In our approach, we focus on the final transformer layer, which is generally believed to carry the most semantically meaningful representations, and to exert the most direct influence on the next token prediction. We compute the attention values by averaging across all attention heads in the last layer. The attention values corresponding to the first two tokens are set to zero, as these correspond to special tokens used to initiate the generation process. For improved interpretability, we apply a series of normalization and filtering steps. First, we compute the mean and standard deviation of the attention weights across the sequence. We then zero out all weights that lie within half a standard deviation of the mean, effectively removing low-signal attention. Finally, to amplify low-to-mid-range attention values and make subtle patterns more visible, we apply a cubic root transformation to the remaining weights.

Figure~\ref{fig:comparison} presents a qualitative comparison of attention weight visualizations in the context of a URL redirection task, analyzed across three prompting frameworks: the baseline, robust prompting, and TypePilot. In the baseline setting, attention is focused on the core task specification, with key terms such as \textit{redirect} receiving prominent attention. The resulting code fulfills the task but remains vulnerable to exploitation. Under the robust prompting framework, the attention distribution shifts slightly, with safety-related tokens in the prompt influencing the generated explanation text. However, these tokens seem to have little effect on the generation of the code itself. On the other hand, in TypePilot there seems to be a clearer effect of the provided vulnerabilities and safety instructions. Tokens corresponding to explicit vulnerabilities in the prompt show a stronger influence on subsequent tokens, suggesting a more grounded incorporation of risk signals. Notably, attention is also concentrated around the instruction to use typed constructs, such as \textit{sealed traits}.

\newpage

% \newpage
% \input{sections/appendix}

\end{document}